\documentclass[runningheads]{llncs}

\usepackage[T1]{fontenc}
\usepackage{graphicx}
\usepackage{booktabs}
\usepackage{array}
\usepackage{amsmath}
\usepackage{amssymb}
\usepackage{xcolor}
\usepackage{url}
\usepackage[hidelinks]{hyperref}
\usepackage{placeins}

\graphicspath{{figures/}}

\begin{document}
\raggedbottom

\title{LISynSeg: Data-Centric Label-to-Image Synthesis for Cross-Modality Whole-Heart Segmentation}
\titlerunning{LISynSeg for Cross-Modality WHS}

\author{Jiacheng Wang\inst{1} \and Ivana Isgum\inst{2} \and Ipek Oguz\inst{1}\thanks{Corresponding author}}
\authorrunning{J. Wang et al.}
\institute{Vanderbilt University, Nashville, TN, USA\\\email{jiacheng.wang.1, ipek.oguz@vanderbilt.edu}\and Mayo Clinic, Rochester, MN, USA\\\email{isgum.ivana@mayo.edu}}

\maketitle

\begin{abstract}
Whole-heart segmentation (WHS) in computed tomography (CT) and magnetic resonance imaging (MRI) is affected by acquisition shifts and heterogeneous cardiac annotations. Existing WHS systems combine architectural design, transfer learning, and generic spatial or intensity augmentation. 
We investigate whether changes to data augmentation and training supervision can improve cross-modality WHS while the segmentation architecture is held constant.
We present LISynSeg, a data-centric approach that augments real-image nnU-Net training with label-to-image synthesis. Synthetic volumes are generated from cardiac label maps using contrast and acquisition perturbations calibrated to the training cohort, then mixed with real images to retain thoracic context absent from the labels (and thus the synthesized images). We model cardiac label variation through controlled changes in myocardial wall thickness and partial supervision of uncertain vessel endpoints. On the CARE Whole-Heart benchmark, synthetic-only training performs worse than the real-image nnU-Net baseline, whereas calibrated real-synthetic training improves cross-modality segmentation without changing the architecture; the improvement is larger for MRI than for CT. The results show that modifying the training data strategy can benefit model development for heterogeneous cardiac data. Code and trained weights will be released at \url{https://github.com/MedICL-VU/Care26_LISynSeg}.

\end{abstract}

\section{Introduction}

\begin{figure}[!t]
    \centering
    \includegraphics[width=\linewidth]{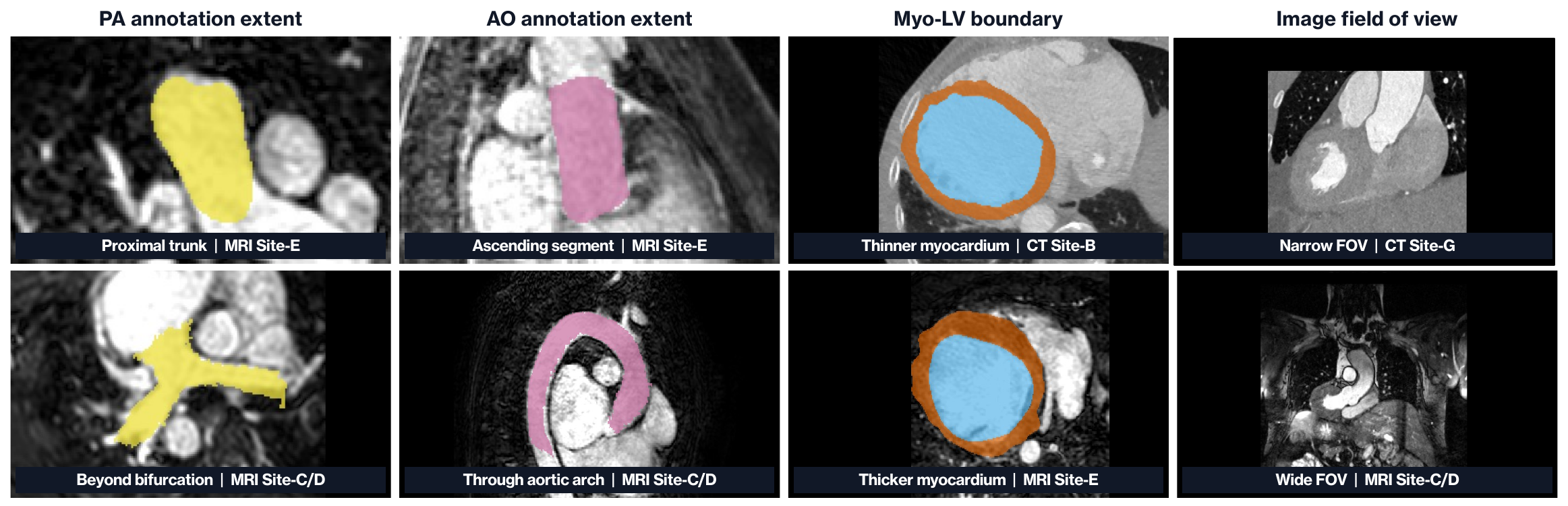}
    \caption{Representative sources of heterogeneity in CARE-WHS CT and MRI. From left to right: pulmonary artery (PA, yellow) annotation limited to the proximal trunk versus extending into both branches beyond the bifurcation; ascending aorta (AO, pink) annotation limited to the ascending segment versus continuing through the arch; thinner versus thicker myocardium at the Myo--LV boundary; and cardiac-focused CT versus wider thoracoabdominal MRI coverage.}
    \label{fig:heterogeneity}
\end{figure}

Whole-heart segmentation (WHS) from computed tomography (CT) and magnetic resonance imaging (MRI) requires a model to delineate seven connected structures: left ventricular myocardium (Myo), left atrium (LA), left ventricle (LV), right atrium (RA), right ventricle (RV), ascending aorta (AO), and pulmonary artery (PA). The CARE Whole-Heart track makes the heterogeneity across scanners, protocols, and centers explicit: images differ in contrast, voxel spacing, orientation, and field of view \cite{care2026_wholeheart,zhuang2019whs}. 
Another source of heterogeneity is variation in the field of view of the scans and in the extent of the manual annotations, which can especially affect the great vessels (AO and PA). 
The CARE challenge defines the AO target as the segment from the aortic valve to the superior level of the atria, and the PA target as the segment from the pulmonary valve to its bifurcation \cite{care2026_wholeheart}.
Owing to the practical difficulty of consistently delineating distal vessel endpoints, the released CARE labels show some variation: the AO annotation can terminate near the proximal ascending segment or continue through the arch, and the PA annotation can end before or after its bifurcation. Thus, predictions for these structures must be evaluated over standardized lengths to account for this variability. 
Figure~\ref{fig:heterogeneity} illustrates representative
variation in scan coverage, AO and PA annotation extent, and Myo--LV boundary
geometry across the released cases.

Most WHS methods address heterogeneity by modifying the segmentation pipeline. Common baselines include 3D U-Net \cite{cicek2016unet3d}, UNETR \cite{hatamizadeh2022unetr}, SwinUNETR \cite{hatamizadeh2022swinunetr}, and nnU-Net \cite{isensee2021nnunet}. Recent CARE submissions have further explored transfer learning with intensity transformations \cite{brosig2026transfer} and implicit-representation modules \cite{zheng2026ieunet}. These approaches modify the model or its training procedure, but they do not explicitly expand the cardiac label geometries or annotation extents represented by the supervision. 
Rather than introducing another segmentation architecture, we investigate whether cross-modality WHS can be improved by changing the images and supervision used during training.

Label-to-image synthesis provides one such intervention while retaining the same segmentation architecture. This strategy has two requirements. First, the synthetic images must span the appearance variation induced by the target acquisitions. We adopt SynthSeg, which samples training images from deformed label maps using class-specific intensity distributions and simulated acquisition effects \cite{billot2023synthseg}. Second, the label maps must represent the anatomical variability of the target task. SIAM addresses this requirement through high-resolution, anatomy-specific label transformations rather than only global affine or elastic deformation \cite{valabregue2026siam}. BabySeg further combines synthetic contrast variation with information from real scans \cite{hoffmann2025babyseg}. Together, these studies motivate using synthesis for appearance variation and designing label operations around the structures that vary in the target task.

CARE-WHS presents two specific difficulties: incomplete context for image synthesis and structure-specific label variability. The seven target classes do not describe surrounding thoracic tissues, so images synthesized from labels alone omit context present in real scans. Meanwhile, label variability differs by structure: myocardial wall thickness changes the Myo--LV boundary, whereas the distal extents of the AO and PA reflect differences in both anatomy and manual labeling practice (Fig.~\ref{fig:heterogeneity}). Global deformation changes pose and overall shape, but it does not directly model either the Myo--LV interface or variable great-vessel endpoints. A cardiac adaptation should therefore use synthesis to complement rather than replace real-image supervision, calibrate appearance variation to observed acquisitions, and apply label operations to the structures that vary in this task.

\begin{figure}[!t]
    \centering
    \includegraphics[width=\linewidth]{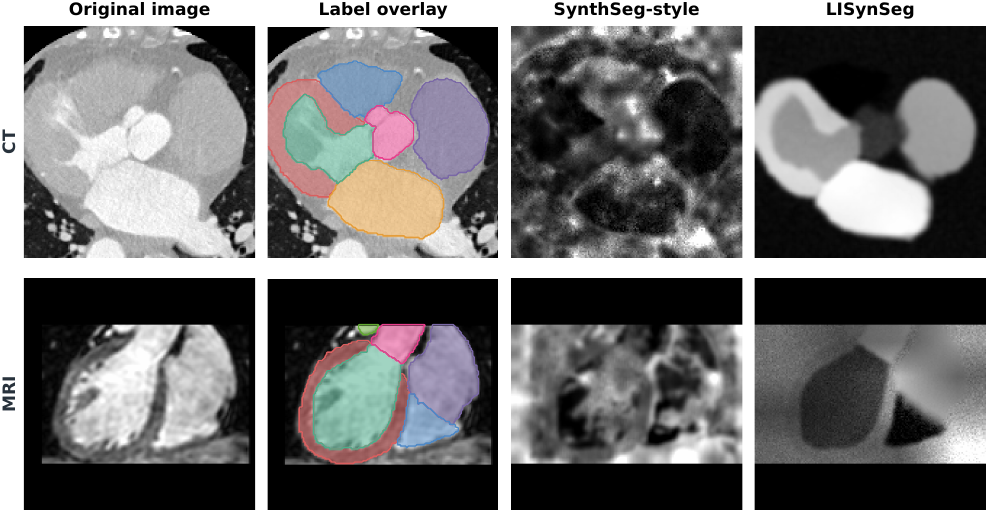}
    \caption{Examples of images generated for CT and MRI. Each row shows a real image, its seven-class label map, an image generated with SynthSeg, and an image generated with LISynSeg. SynthSeg uses generation regions obtained from real images, whereas LISynSeg uses the seven cardiac label regions and parameters calibrated from the training data. Spatial transformations are omitted from this comparison.}
    \label{fig:synthesis}
\end{figure}

LISynSeg follows these design choices. Real CT and MRI examples are mixed with images synthesized from labels using cohort-calibrated acquisition parameters. Shape variation is introduced at the Myo--LV boundary, and supervision is relaxed in distal vessel regions whose annotated extent varies across cases. Figure~\ref{fig:synthesis} contrasts images generated by LISynSeg and a SynthSeg-style baseline from the same CT and MRI label maps. The controlled comparisons use a common segmentation architecture and inference pipeline to isolate the effect of training data construction. 
Separately, we report the result of our submitted ensemble in the official CARE 2026 validation phase, where it ranked top five on the combined CT and MRI leaderboard at the time of writing.

Our main contributions are as follows:
\begin{itemize}
    \item We investigate whether cross-modality WHS can be improved by changing the training data and supervision while keeping the segmentation architecture and inference procedure fixed.
    \item We propose LISynSeg, a real--synthetic training framework that combines cohort-calibrated image synthesis with Myo--LV boundary transformations and partial supervision at variable distal vessel endpoints.
    \item We show that real--synthetic training improves over real-only and synthetic-only baselines, particularly on the more challenging MRI cases.
\end{itemize}

\section{Method}

\begin{figure}[!t]
    \centering
    \includegraphics[width=\linewidth]{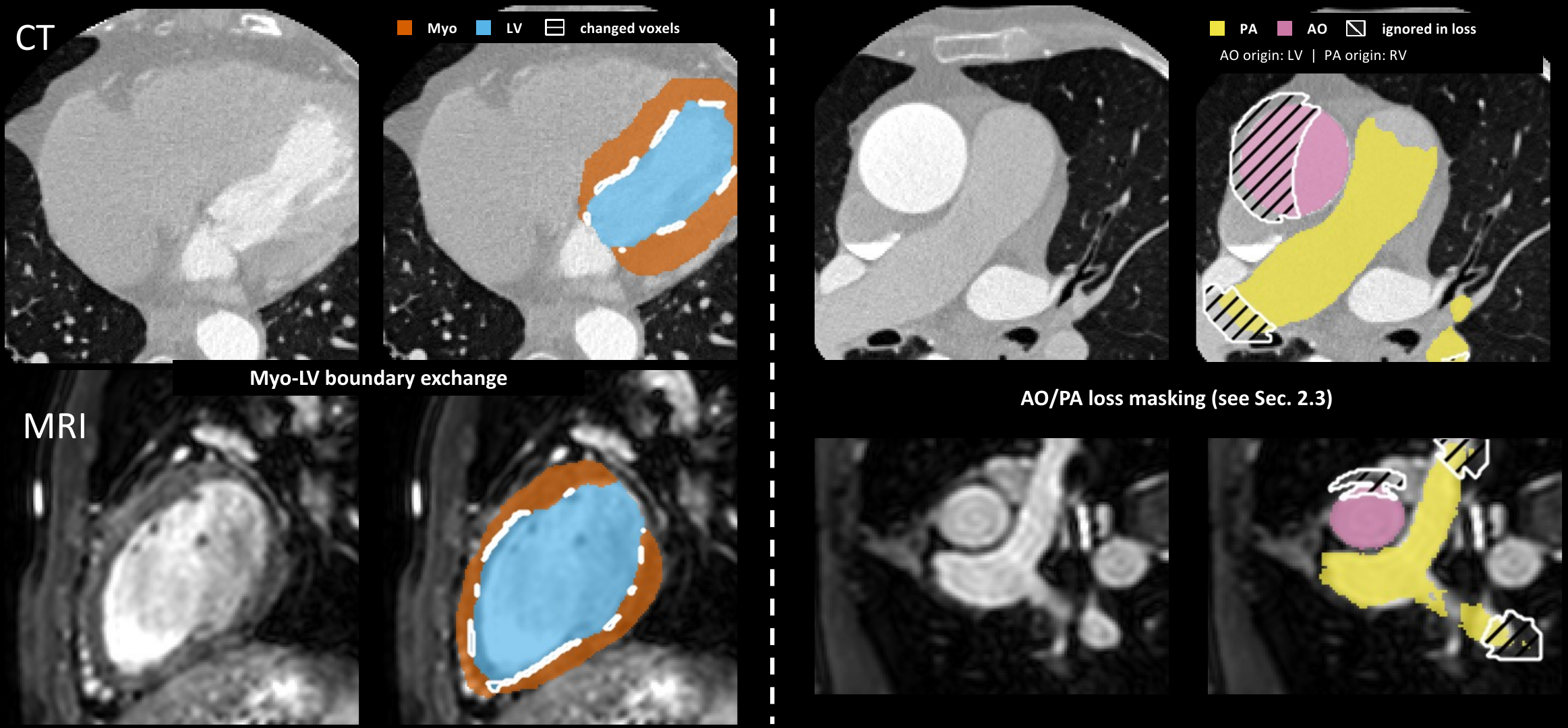}
    \caption{Cardiac-specific operations in LISynSeg. Left: Myo--LV boundary exchange thickens or thins the myocardium by moving the endocardial interface while preserving the combined Myo/LV extent; white contours denote exchanged voxels. Right: AO/PA loss masking retains proximal vessel supervision while excluding distal voxels whose annotated extent varies across cases; hatching marks voxels ignored by the segmentation loss. The proximal side is identified from AO--LV and PA--RV adjacency (Sec.~2.3). CT and MRI rows show both operations across modalities.}
    \label{fig:cardiac_design}
\end{figure}

LISynSeg retains a standard 3D full-resolution nnU-Net as the segmentation backbone and changes only the construction of training examples and supervision. We first describe how label-to-image synthesis is incorporated into real-image training, then introduce two cardiac-specific operations for the Myo--LV interface and distal great-vessel endpoints (AO and PA). Figure~\ref{fig:cardiac_design} illustrates these operations in CT and MRI. The network architecture, preprocessing, optimization, and inference procedure remain fixed across the controlled comparisons.

\subsection{Label-to-Image Synthesis as Training Augmentation}

SynthSeg trains a segmentation network from label maps by applying spatial deformation and generating images with randomly sampled class intensities and acquisition effects \cite{billot2023synthseg}. Billot et al. evaluated this approach on the Multi-Modality Whole Heart Segmentation (MMWHS) challenge, which preceded CARE-WHS. Their experiment used 13 MRI label maps containing the same seven cardiac structures considered here. Because these labels did not represent all visible tissues, intensities from the corresponding MRI scans were clustered to divide each foreground label into two generation regions and the background into three to ten regions. These regions controlled image synthesis, while the segmentation target remained the original seven-class map \cite{billot2023synthseg,zhuang2019whs}.

LISynSeg uses label-to-image synthesis to augment rather than replace real-image training. Let $(x_i^a,y_i^a)=D(x_i,y_i)$ denote a real image and label after standard nnU-Net augmentation, and let $y_i^g=G(y_i^a;\eta_i)$ denote the label after the Myo--LV operation described in Sec.~2.2. For each training example, we sample $z_i\sim\mathrm{Bernoulli}(\rho)$ with $\rho=0.10$ and use
\begin{equation}
(x_i',y_i')=
\begin{cases}
\left(S(y_i^g;\theta_i),y_i^g\right), & z_i=1,\\
\left(x_i^a,y_i^a\right), & z_i=0.
\end{cases}
\label{eq:mixture}
\end{equation}
Here, $z_i=1$ selects the synthetic branch, in which $y_i^g$ is used both to synthesize the image and as the segmentation target, whereas $z_i=0$ retains the augmented real image--label pair.
For the synthesis ablation without boundary exchange, $G$ is the identity transformation. We do not apply an additional affine or elastic deformation within $S$ because nnU-Net has already spatially transformed the image and label.

The synthesis function operates directly on the seven cardiac labels and the background. For every label value present in the map, its Gaussian mean and standard deviation are sampled independently as $\mu_l\sim U(0,255)$ and $\sigma_l\sim U(0,12)$. We then simulate partial-volume effects, a bias field with standard deviation sampled from $U(0,0.35)$, gamma variation with $\log\gamma\sim\mathcal{N}(0,0.15^2)$, and Gaussian noise with standard deviation sampled from $U(0,3)$. Resolution profiles are sampled from the native-to-target spacing ratios observed among training subjects in the current fold and are capped at a factor of three. Each synthesized image is finally standardized to zero mean and unit variance.

\subsection{Myo--LV Boundary Exchange}

Myocardial thickness, determined by the Myo and LV boundaries, is one of the
major sources of heterogeneity in the WHS datasets. Generic deformation models
provide little direct control over this thickness; we therefore introduce an
explicit Myo--LV boundary exchange step.

Let $M_i$ and $V_i$ denote the Myo and LV voxel sets in the augmented label
map $y_i^a$. For any binary voxel set $Q$, let $\mathcal{D}_1(Q)$ denote its
dilation using a $3\times3\times3$ neighborhood. The two interface bands are
$B_{V\rightarrow M}=\mathcal{D}_1(M_i)\cap V_i$ and
$B_{M\rightarrow V}=\mathcal{D}_1(V_i)\cap M_i$.

For each selected synthetic example, thickening or thinning is sampled with
equal probability: thickening transfers voxels from $B_{V\rightarrow M}$ to
Myo, whereas thinning transfers voxels from $B_{M\rightarrow V}$ to LV. The left panels of Fig.~\ref{fig:cardiac_design} show examples of this operation in CT and MRI, with the exchanged voxels indicated by white contours.
Rather than transferring the complete interface band, we select a spatially
smooth subset using a random field; the selected voxels are not required to
form a single connected component.

The largest admissible exchange is determined by requiring the Myo and LV
volume ratios to remain within $[0.72,1.35]$ and $[0.80,1.20]$, respectively,
and the number of exchanged voxels is sampled as a fraction
$\lambda\sim\mathcal{U}(0.35,0.85)$ of this limit. The transformation preserves
$M_i^g\cup V_i^g=M_i\cup V_i$ and leaves $G(y_i^a;\eta_i)(v)=y_i^a(v)$ for
$v\notin M_i\cup V_i$. If either class is absent from the sampled patch or no
exchange satisfies the volume constraints, $G$ returns the original label map.
This boundary exchange defines $G$ in Eq.~\ref{eq:mixture} and is applied only
to examples selected for synthesis.

\subsection{Distal Great Vessel Loss Masking}

The CARE protocol defines the intended AO and PA extents, but the released
annotations vary at distal endpoints that are not always marked by visible
anatomical boundaries. Standard Dice and cross-entropy losses nevertheless treat
every voxel as a fixed target, including voxels near these endpoints; this can
produce conflicting supervision when annotations encode different endpoint
extents.

For vessel $c\in\{\mathrm{AO},\mathrm{PA}\}$, we associate the AO with the LV
and the PA with the RV. Using the dilation operator from Sec.~2.2, the proximal
region $R_c$ contains vessel voxels in contact with
$\mathcal{D}_1(\mathrm{LV})$ for the AO or
$\mathcal{D}_1(\mathrm{RV})$ for the PA. If no ventricular contact is present
in the sampled patch, no endpoint mask is applied. The right panels of Fig.~\ref{fig:cardiac_design} show the corresponding
ventricle used to locate the proximal vessel region and the distal voxels
excluded from the segmentation loss.

Let $\mathbf{s}$ denote the nnU-Net target spacing and
$p(v)=\mathbf{s}\odot v$ the physical coordinate of voxel $v$. The proximal
reference point is the mean coordinate
$r_c=|R_c|^{-1}\sum_{v\in R_c}p(v)$, and the distance of vessel voxel $v$ from
this point is $d_c(v)=\lVert p(v)-r_c\rVert_2$.

The closest 90\% of annotated vessel voxels retain their original supervision,
whereas the remaining distal voxels are assigned to the ignore set $U_c$. The
distal cap is defined by the 96th percentile of $d_c$, and $U_c$ additionally
includes background within a four-voxel dilation of this cap when its distance
from $r_c$ is at least the cap distance, allowing a tolerance of half the
smallest voxel spacing. If tied distance values cause the distal selection to
exceed 40\% of the vessel, masking is skipped.

For either branch of Eq.~\ref{eq:mixture}, the mask is computed from the
resulting target $y_i'$ immediately before loss computation. Writing
$U_i=U_{\mathrm{AO}}\cup U_{\mathrm{PA}}$ and denoting the segmentation network
by $f_\phi$, the objective is
\begin{equation}
    \mathcal{L}_{\mathrm{seg}}
    =
    \mathcal{L}_{\mathrm{Dice+CE}}
    \left(f_\phi(x_i'),y_i';\Omega_i\setminus U_i\right).
    \label{eq:partial}
\end{equation}
The operation changes which voxels contribute to the loss without generating
an extended or truncated vessel label. It is applied to every training example,
regardless of whether the input is real or synthetic, and is not used during
validation or inference. The masking rule therefore does not explicitly favor
either extension or truncation of the annotated vessel endpoint.

\subsection{Training and Inference}

Each training batch first undergoes the spatial and intensity augmentations used by nnU-Net. For an example selected with probability $\rho=0.10$, the Myo--LV operation is applied to the transformed label map, and the augmented real image is replaced by an image synthesized from the resulting map. In the complete method, the AO and PA ignore regions defined in Sec.~2.3 are excluded from the Dice and cross-entropy losses immediately before loss computation. For each selected example, we independently sample the class intensity distributions, bias field, gamma transformation, noise level, and resolution profile described in Sec.~2.1, as well as whether the myocardium is thickened or thinned. At inference, the network receives one preprocessed image channel without a modality indicator or synthesis step. The same 3D full-resolution configuration is used for CT and MRI; no 2D model or slice-wise inference is used. We otherwise use the standard nnU-Net inference procedure.

\section{Experiments and Results}

\subsection{Data, Training, and Evaluation Protocol}

The released CARE-WHS training set contains 106 labeled volumes from five site groups: 60 CT and 46 MRI scans. All images and annotations used in this study were released through the CARE 2026 Whole Heart track \cite{care2026_wholeheart,zhuang2016atlas,zhuang2019mvm,gao2023bayeseg}; no private images or annotations were used. Every model is trained with both CT and MRI cases, and the two modalities are separated only when validation results are reported. We use the same partition into five folds as the nnU-Net baseline. Tables~\ref{tab:main} and~\ref{tab:components} report one development split containing 85 training cases and 21 validation cases, of which 12 are CT and 9 are MRI. With the exception of the ``Only synthetic'' experiment listed in Table~\ref{tab:main}, the nnU-Net comparisons use the same augmentation, optimization schedule, and 200-epoch budget. The 3D U-Net, UNETR, and SwinUNETR baselines use the same development split, preprocessed inputs, and evaluation procedure.

For the model comparisons, we report the Dice similarity coefficient (DSC) separately for CT and MRI. Within each reported group, DSC is first averaged across cases for each of the seven structures. The overall WHS score is the weighted mean of these seven structure averages, using the total reference-annotation volume of each structure as its weight. Each label in each model is postprocessed by extracting its largest connected component before evaluation.

\subsection{Segmentation Model and Training Strategy Comparison}

Table~\ref{tab:main} compares four segmentation models trained on only real
images and four training settings based on nnU-Net.

\begin{table}[!b]
\centering
\caption{Comparison of segmentation models and training strategies on the
development split. All models are trained jointly on CT and MRI; CT and MRI
report results on the corresponding validation subsets, and All reports the
combined score. \textbf{Bold} and \underline{underlining} indicate the highest and second-highest DSC, respectively.}
\label{tab:main}
\small
\setlength{\tabcolsep}{3.5pt}
\begin{tabular}{@{}>{\raggedright\arraybackslash}p{0.35\linewidth}
                    >{\raggedright\arraybackslash}p{0.25\linewidth}ccc@{}}
\toprule
Segmentation model & Training setting & CT & MRI & All \\
\midrule
3D U-Net\cite{cicek2016unet3d}
    & Only real & .8723 & .7891 & .8215 \\
UNETR\cite{hatamizadeh2022unetr}
    & Only real & .9123 & .8007 & .8432 \\
SwinUNETR\cite{hatamizadeh2022swinunetr}
    & Only real & .9011 & .7685 & .8337 \\
nnU-Net\cite{isensee2021nnunet}
    & Only real & \underline{.9517} & .8890 & .9203 \\
\midrule
nnU-Net\cite{isensee2021nnunet}
    & Only synthetic & .9119 & .8303 & .8735 \\
nnU-Net\cite{isensee2021nnunet}
    & Real and synthetic & .9503 & \underline{.8912}
    & \underline{.9208} \\
nnU-Net\cite{isensee2021nnunet}
    & Full LISynSeg
    & \textbf{.9519} & \textbf{.8917} & \textbf{.9219} \\
\bottomrule
\end{tabular}
\end{table}

Among the four models trained on only real images,
nnU-Net~\cite{isensee2021nnunet} achieved the highest DSC on this development
split. The two Transformer models, UNETR~\cite{hatamizadeh2022unetr} and
SwinUNETR~\cite{hatamizadeh2022swinunetr}, did not outperform nnU-Net under
the evaluated settings. These results compare the specific implementations
tested here and do not establish a general advantage of one architecture type
over another.

Training nnU-Net with only synthetic images performed worse than training with
only real images, with the largest decrease observed on MRI
($-5.87$ percentage points). This comparison shows that synthetic images
cannot replace real images under the current setting and are instead used
alongside them during training. Using real and synthetic images increased MRI and overall DSC by 0.22 and 0.05 percentage points, respectively, while CT DSC decreased by 0.14 percentage points. The full LISynSeg setting, which additionally includes the two cardiac label operations, achieved the highest DSC in all three columns. These gains were obtained without making the segmentation model larger or changing the inference procedure.

\subsection{Ablation of Cardiac Label Operations}

Table~\ref{tab:components} evaluates the Myo--LV boundary operation and AO/PA
loss masking individually and together using the same base setting with real
and synthetic images.

\begin{table}[!b]
\centering
\caption{Ablation of the cardiac label operations using the same base setting
with real and synthetic images. All models are trained jointly on CT and MRI
using the same development split, training schedule, synthesis probability,
and postprocessing. The CT, MRI, and All columns report aggregate scores across
all seven structures. \textbf{Bold} and \underline{underlining} indicate the highest and second-highest DSC, respectively.}
\label{tab:components}
\small
\setlength{\tabcolsep}{5.0pt}
\begin{tabular}{@{}lccccc@{}}
\toprule
Training setting & Myo--LV & AO/PA & CT & MRI & All \\
\midrule
Real and synthetic
    & -- & -- & .9503 & \underline{.8912} & \underline{.9208} \\
\quad + Myo--LV boundary operation
    & \checkmark & -- & .9510 & .8672 & .9147 \\
\quad + AO/PA loss masking
    & -- & \checkmark & \underline{.9513} & .8901 & .9207 \\
\quad + both operations
    & \checkmark & \checkmark
    & \textbf{.9519} & \textbf{.8917} & \textbf{.9219} \\
\bottomrule
\end{tabular}
\end{table}

The Myo--LV boundary operation alone produced a small increase on CT but reduced MRI DSC by 2.40 percentage points and also lowered the overall DSC. The operation therefore produced different responses on CT and MRI. AO/PA loss masking alone remained close to the base setting across all three columns. When both operations were enabled, the model achieved the highest numerical DSC on CT, MRI, and overall. Although the differences from the base setting were small, they were positive in all three columns.

\begin{figure}[!t]
    \centering
    \includegraphics[width=\linewidth]{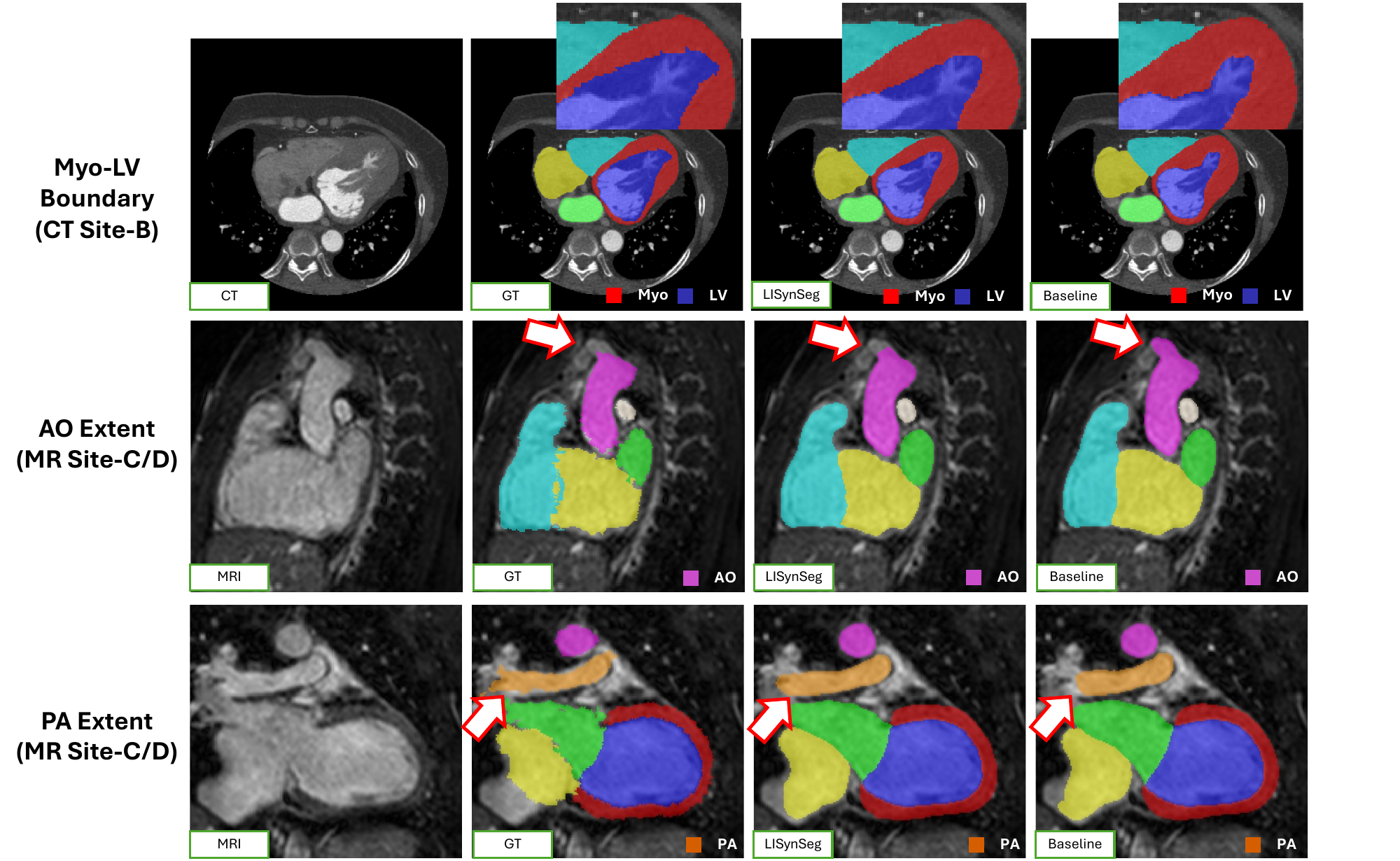}
    \caption{Qualitative comparison on representative cases. Columns show the input image, ground truth, LISynSeg, and the real-image nnU-Net baseline. From top to bottom: a CT example at the Myo--LV interface and MRI examples with the annotated AO (middle row) and PA (bottom row) extents. The zoomed-in panel in the top row shows a more accurate Myo--LV interface produced by our method. Red arrows in the bottom two rows mark the distal vessel differences.}
    \label{fig:qualitative}
\end{figure}

\subsection{Qualitative Comparison}

Although the quantitative results summarize overall performance, they do not fully reveal whether LISynSeg mitigates the specific segmentation discrepancies targeted by our design. Figure~\ref{fig:qualitative} therefore visually compares the real-image nnU-Net baseline with LISynSeg on one CT case and two MRI cases from the validation set.

For the CT Site-B case (top row), LISynSeg more closely follows the endocardial Myo--LV interface and reduces the LV under-segmentation observed in the baseline. Accordingly, the LV Dice score increases from 0.83 to 0.86, while the Myo Dice score increases from 0.75 to 0.79 for this subject. 
In the MRI Site-C and Site-D cases, LISynSeg produces segmentations that are more consistent with the released ground-truth annotations. Specifically, it captures less of the AO annotation in the Site-D case and a larger portion of the PA annotation in the Site-C case, thereby reducing the AO over-segmentation and PA under-segmentation observed in the baseline. In the middle row, the AO Dice score increases from 0.73 to 0.75, while in the bottom row, the PA Dice score increases from 0.67 to 0.71.

These examples localize the improvements to the anatomical structures targeted by the cardiac-specific label operations and complement the aggregate quantitative results reported in Table~\ref{tab:components}. Because the annotated endpoints of the great vessels vary across the released labels, the AO and PA panels illustrate agreement with the released annotations rather than the standardized vessel-length evaluation used by the challenge.

\subsection{Official Validation Ensemble}

Our official validation submission combines nnU-Net models trained with only real images, with real and synthetic images, and with the data augmentation (DA5) setting. In the public validation leaderboard, the ensemble achieved DSCs of 0.9360 for CT, 0.8640 for MRI, and 0.9072 for the combined track. The submission ranked among the top five on the combined track at the time of writing. These results report the performance of the submitted ensemble, whereas Tables~\ref{tab:main} and~\ref{tab:components} isolate the contribution of the data augmentation strategy.

\section{Discussion}

LISynSeg improves WHS across CT and MRI by combining label-to-image synthesis calibrated to the training cohort with cardiac label augmentation and loss masking, without modifying the segmentation architecture. The improvement is larger for MRI than for CT, suggesting that the added contrast and acquisition variability is particularly useful for MRI. Improvements are also observed for the great vessels. The second and third rows of Fig.~\ref{fig:qualitative} show closer agreement with the reference AO and PA segmentations, particularly around vessel branches and distal extents.

Our approach extends the central idea of SynthSeg, whose cardiac experiment on MMWHS, a predecessor of CARE-WHS, showed that randomized images generated from label maps can support segmentation across CT and MRI \cite{billot2023synthseg,zhuang2019whs}. CARE-WHS introduces greater variation across centers in scan coverage and in the annotated extent of the AO and PA. Since the distal vessel endpoints do not correspond to a consistently visible anatomical boundary, image randomization alone cannot account for all variation in the training labels. We therefore combine synthesized examples with real images and directly account for variation in the cardiac annotations. Mixing real and synthetic examples broadens the contrast and acquisition conditions observed during training while retaining thoracic tissues that are not represented by the seven labels. The Myo--LV boundary exchange and AO/PA endpoint loss masking address the two challenges described above: the former introduces controlled variation in myocardial thickness, whereas the latter reduces the influence of inconsistent distal vessel annotations during optimization. Together, these findings show that cardiac label-to-image synthesis benefits from combining image randomization with targeted treatment of annotation variation.

\section{Conclusion}

LISynSeg expands real CT/MRI training with calibrated label-to-image synthesis while holding the nnU-Net architecture fixed. Our experiments show that real--synthetic training improves MRI and overall segmentation over nnU-Net trained using only real images, whereas synthetic-only training loses image context that is absent from the labels. Cardiac boundary augmentation and AO/PA loss masking address two sources of label heterogeneity that generic deformation does not. These results identify training data augmentation design as a practical route for cross-modality WHS.

\begin{credits}
\subsubsection{\ackname} We thank the Advanced Computing Center for Research and Education (ACCRE) at Vanderbilt University for providing computational resources. We also thank the CARE Whole-Heart (CARE-WHS) Challenge organizers for providing the data used in this study. No external datasets were used.
\subsubsection{\discintname} The authors have no competing interests to declare that are relevant to the content of this article.
\end{credits}

\bibliographystyle{splncs04}
\bibliography{references}

\end{document}